\documentclass{article}
\usepackage{iclr2027_conference,times}
\iclrfinalcopy
\usepackage{graphicx}
\usepackage{amsmath,amssymb}
\usepackage{booktabs}
\usepackage{placeins}
\usepackage{hyperref}
\usepackage{xurl}
\graphicspath{{figures/}}
\newcommand{\method}{STR}
\newcommand{\ASR}{\operatorname{ASR}}
\newcommand{\KL}{\operatorname{KL}}

\title{\method: Supervised Transcoder Replacement\\for Reducing Steering Side Effects}
\author{
 \textbf{Haonan Yu\textsuperscript{1}},
 \textbf{Junhao Liu\textsuperscript{2}},
 \textbf{Zhenyu Yan\textsuperscript{3}},
 \textbf{Haoran Lin\textsuperscript{4}},
\\
 \textbf{Xin Zhang* \textsuperscript{5}}}
\date{September 2026}

\begin{document}

\maketitle

\begin{abstract}
Model steering can strengthen a target behavior while degrading other useful behaviors. We introduce Supervised Transcoder Replacement (\method) to reduce these side effects for existing steering methods, including those fitted without a protection objective. \method\ learns a replacement for the multilayer perceptron (MLP) computation at the steering layer through supervision for target control, non-target preservation, and fidelity without steering. Selected steering methods then fit directions on the frozen replacement while retaining their own fitting objectives. We evaluate three steering methods across Gemma and Llama models using Corrigibility preferences and four harmful-request safety datasets. SALAD-Bench supplies protection training data and a separate in-distribution evaluation split; HarmBench, AdvBench, and StrongREJECT are reserved for out-of-distribution testing. STR substantially reduces steering side effects on the in-distribution evaluation and extends this protection to the unseen safety datasets while retaining effective target control. For target-only supervised steering vectors, pooled out-of-distribution attack success rate falls from 42.46\% to 14.42\% on Gemma-3-4B and from 34.97\% to 12.91\% on Gemma-3-12B. These results show that replacement training can benefit steering methods fitted without protection objectives.
\end{abstract}

\section{Introduction}
\label{sec:introduction}

Model steering provides a flexible way to change a language model's behavior through interventions on its internal representations \citep{turner2023activation,li2023iti,zou2023repe}. A useful intervention should strengthen the desired behavior while preserving other behaviors and capabilities that make the model useful. However, existing steering methods can introduce side effects, such as weakened safety behavior and increased vulnerability to jailbreaks \citep{aravindan2026opium,xiong2026externalities,goyal2026specificity}. Reducing these side effects while retaining effective target control is therefore a central challenge for model steering.

Existing steering procedures differ in how they obtain a direction and which behaviors their objectives address. Some use target examples alone \citep{rimsky2024caa,yin2024lofit}, while others explicitly incorporate preservation requirements \citep{aravindan2026opium}. A common setup applies the intervention at a single selected layer \citep{turner2023activation,rimsky2024caa}, although multi-layer interventions are also used \citep{zou2023repe}. We focus on this single-layer setting.

Our approach is motivated by representational overlap: target and non-target features can share the same coordinates. Individual neurons can respond to multiple concepts, a phenomenon known as polysemanticity and commonly linked to feature superposition \citep{cunningham2023sae}. An intervention that changes coordinates used by a target feature can therefore also perturb computations supporting other behaviors. Transcoders approximate a densely activated MLP through a wider, sparsely activated intermediate representation \citep{dunefsky2024transcoders}. This suggests a route to reducing interference: replace the computation at the steering layer with a sparse transcoder and train it to support target control while preserving specified non-target behaviors.

We propose Supervised Transcoder Replacement (\method) to implement this approach. We train a transcoder to replace one MLP mapping at the steering layer, providing a sparse feature space for intervention. Inspired by auxiliary supervision \citep{lee2015deeplysupervised}, we introduce an auxiliary vector $u$ that creates steering interventions during training. We jointly optimize $u$ and the transcoder while keeping the surrounding language model frozen. These interventions let training directly account for what happens when the replacement is steered: supervision encourages stronger target behavior and preservation of specified non-target behavior under intervention, alongside fidelity to the original model when control is disabled. After training, we freeze the replacement, set aside $u$, and fit existing steering methods on the resulting model.

Then the steering method fits its direction in the replacement's representation space using its original fitting rule and control objective. Because preservation is learned during replacement training, a downstream method can benefit without adding a protection term to its objective. 

We evaluate contrastive activation addition (CAA), target-only supervised steering vectors (SSV), and protection-aware OPIUM on the original and replacement versions of Gemma-3-4B-it, Llama-3.1-8B-Instruct, and Gemma-3-12B-it \citep{rimsky2024caa,aravindan2026opium,gemmateam2025gemma3,grattafiori2024llama3}. Target control is measured using Corrigibility choice preferences from Anthropic's model-written evaluations \citep{perez2022modelwritten}, alongside harmful-request safety measured by attack success rate (ASR). SALAD-Bench \citep{li2024saladbench} is the sole source of protection training and safety development data. HarmBench \citep{mazeika2024harmbench}, AdvBench \citep{zou2023universal}, and StrongREJECT \citep{souly2024strongreject} are excluded from training and development and used to test whether the learned protection generalizes to unseen safety datasets.

The central result is that steering methods without their own protection objectives become substantially less disruptive on the STR replacement. SSV shows the most consistent pattern across models: on the pooled unseen safety datasets, its ASR falls from 34.97\% to 12.91\% on Gemma-3-12B, accompanied by higher target utility. On Llama-3.1-8B, target utility rises from 74.10\% to 84.73\%, while ASR also decreases. On Gemma-3-4B, SSV sharply reduces ASR with little observed change in target utility. CAA likewise lowers ASR across all three models, although its utility trade-off varies. These improvements extend the protection observed on SALAD to unseen safety sources, even though neither CAA nor SSV adds a protection term to its fitting objective. OPIUM also benefits more from replacement on Gemma than on Llama.

Our contributions are: (1) supervised transcoder training that provides a behavior-preserving replacement at the steering layer; (2) a two-stage design that supplies protection through the replacement while preserving existing steering methods' fitting objectives; and (3) comparisons across three steering methods, three models, and four safety datasets, showing that protection learned from SALAD transfers to unseen safety sources while retaining effective target control.

\section{Background}
\label{sec:background}

This section reviews activation steering, target control and non-target preservation, and sparse transcoder replacements as background for \method.

\subsection{Activation steering}
\label{sec:activationsteering}
For a model $M$, activation steering modifies an internal representation using a direction and an intervention strength. For the vector-based methods considered here, let $j$ index the selected steering method, $\operatorname{Fit}_j$ denote its direction-fitting rule at a chosen representation site, and $\mathcal D_j$ denote the examples and supervision required by that rule. Let $v_j$ be the fitted direction, $\alpha$ the steering strength, and $h$ and $h'$ the representations before and after intervention, respectively. The intervention operator $\operatorname{Steer}$ maps the original representation and scaled direction to the modified representation. We write the fitting and application steps as
\begin{equation}
v_j=\operatorname{Fit}_j(M,\mathcal D_j),
\qquad h'=\operatorname{Steer}(h,\alpha v_j),
\label{eq:steeringmethod}
\end{equation}
For a generic steering direction $v$, the common additive form is $\operatorname{Steer}(h,\alpha v)=h+\alpha v$ \citep{turner2023activation}. The fitting rule can estimate a direction from representation statistics \citep{rimsky2024caa} or optimize a behavioral objective \citep{yin2024lofit,he2025saessv}; it uses the representations and intervention rule of the model on which it is applied.

A controller consists of a representation site, a fitted direction, and an intervention strength. For a direction $v$, we denote enabled control by $(M,\alpha v)$ and disabled control by $(M,0)$. Target utility $U(M,\alpha v)$ measures performance on the desired behavior. The fitting rule and the model computation receiving the direction jointly determine the intervention's effects.

\subsection{Target control and non-target preservation}
Let $A$ be the desired behavior and $B$ a non-target behavior to preserve. A steering side effect occurs when a procedure intended to improve $A$ degrades performance on $B$ \citep{aravindan2026opium,goyal2026specificity}. Write $E_B(M,\alpha v)$ for a measure of degradation on $B$, with larger values indicating worse outcomes. Effective control increases $U$ while keeping $E_B$ low.

\subsection{Transcoders as layer replacements}
A sparse autoencoder reconstructs the activation that it encodes \citep{cunningham2023sae}. A transcoder instead learns a sparse input--output approximation to a computation such as an MLP mapping \citep{dunefsky2024transcoders}. Its encoder maps inputs to latent coordinates; its decoder maps them to the replaced computation's output space. The decoded activation replaces that computation's output in the forward pass. This provides both a replacement computation and a representation space in which existing steering rules can be applied. A single MLP replacement preserves the surrounding attention, residual connections, and other layers. \method\ trains the replacement for fidelity on unsteered inputs and behavior preservation during steering.

\section{Method}
\label{sec:method}

\method\ reduces steering side effects by learning the computation that receives the intervention. Figure~\ref{fig:workflow} shows two stages. In stage (A), we construct a sparse transcoder replacement and jointly train it with an auxiliary steering vector, using supervision for target control, protection under intervention, and fidelity without intervention (Section~\ref{sec:replacementtraining}). In stage (B), we freeze the replacement, fit the selected method's latent direction, and apply it at a chosen strength (Section~\ref{sec:deployment}).

\subsection{Training the replacement}
\label{sec:replacementtraining}
\label{sec:objective}
\begin{figure}[hbp]
\centering
\includegraphics[width=\linewidth]{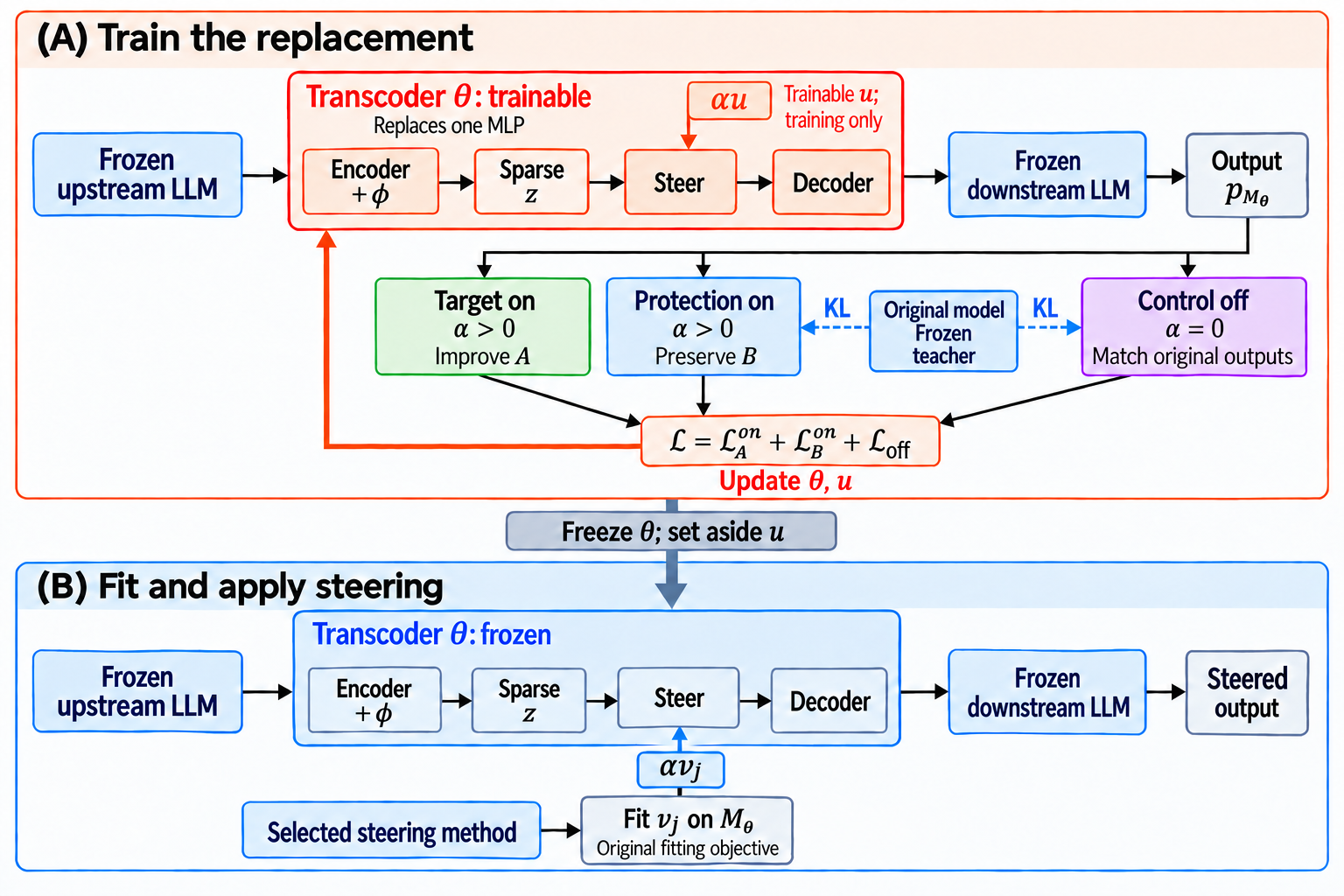}
\caption{Overview of \method. \textbf{(A) Train the replacement.} A transcoder replaces one MLP mapping and is trained with an auxiliary vector $u$ using three objectives: improve target behavior under steering, preserve protected behavior under the same steering, and preserve unsteered outputs. The original model $M_0$ provides the teacher distributions; the surrounding language model remains frozen. \textbf{(B) Fit and apply steering.} After freezing the transcoder and setting aside $u$, a selected steering method fits a latent direction $v_j$ under its original objective and applies it with strength $\alpha$. Both JumpReLU and TopK follow this workflow, using their respective intervention rules in Equation~\ref{eq:intervention}.}
\label{fig:workflow}
\end{figure}

This stage follows Figure~\ref{fig:workflow}(A): install the replacement, construct training-time interventions, evaluate their behavioral consequences, and update the trainable parameters.

\paragraph{Construct the replacement model.}
Let $M_0$ be the original language model, which also serves as the teacher, and let $\theta$ denote the trainable parameters of a transcoder. The replacement model $M_\theta$ is obtained by substituting this transcoder for one MLP mapping. The surrounding model stays frozen.

The transcoder encodes the input activation of the replaced MLP mapping into a sparse latent representation and decodes it back to that mapping's output space. The decoded activation replaces the original mapping's output in the model's forward pass. Let $x$ be the input activation, $z$ the sparse latent activation, and $y$ the decoded output activation, all written as row vectors. The encoder weight matrix $W_{\mathrm{enc}}$ maps the input to latent coordinates, and its bias vector $b_{\mathrm{enc}}$ provides an additive offset. The decoder weight matrix $W_{\mathrm{dec}}$ maps latent coordinates back to the MLP output space, with output bias $b_{\mathrm{dec}}$. These four parameters form $\theta$. Let $\phi$ denote the sparsifying activation function applied after the encoder's affine mapping. The unsteered computation is
\begin{equation}
z=\phi(xW_{\mathrm{enc}}+b_{\mathrm{enc}}),
\qquad
y=zW_{\mathrm{dec}}+b_{\mathrm{dec}}.
\label{eq:transcoder}
\end{equation}
Our implementations use JumpReLU \citep{rajamanoharan2024jumprelu} for the Gemma transcoders and TopK for the Llama transcoder. JumpReLU produces nonnegative latent activations, whereas the TopK implementation retains a fixed number of encoder activations that can include negative values.

\paragraph{Introduce the auxiliary steering intervention.}
Inspired by auxiliary supervision, which uses additional training objectives to shape learned representations \citep{lee2015deeplysupervised}, we introduce $u$ to make the replacement's response to steering part of its training objective. The vector $u$ is a learnable control variable that constructs actual steering perturbations in the transcoder's sparse feature space. Joint optimization learns a target-directed intervention while shaping the replacement's response to it. The auxiliary vector therefore makes training intervention-aware: the transcoder learns from the behavioral consequences of perturbing its features, including effects on protected behavior.

The intervention acts on a support $I$, the index set of latent coordinates eligible for steering, with size $K=|I|$. Its complement $\bar I$ contains all remaining latent coordinates, which receive no intervention. A subscript $I$ or $\bar I$ selects the corresponding entries of a vector. We keep $I$ fixed during replacement training and subsequent controller fitting; Appendix~\ref{sec:implementation} describes its selection. Separately, $L_0$ denotes the number of nonzero entries in the encoded activation $z$, which can vary across inputs.

The intervention occurs after the encoder nonlinearity and before decoding. Let $z'$ be the intervened latent activation and $y'$ its decoded output. For a latent direction $v$, we use an intervention rule that matches the latent activation structure: nonnegative clipping for our JumpReLU transcoders and signed addition for our TopK transcoder:
\begin{equation}
\begin{gathered}
z'_I=\begin{cases}
\max(z_I+\alpha v_I,0), & \text{JumpReLU},\\
z_I+\alpha v_I, & \text{TopK},
\end{cases}\\
z'_{\bar I}=z_{\bar I},\qquad y'=z'W_{\mathrm{dec}}+b_{\mathrm{dec}}.
\end{gathered}
\label{eq:intervention}
\end{equation}
The maximum is applied elementwise. In the TopK case, we decode the intervened activations directly, without applying TopK a second time.
During replacement training, $v=u$ in Equation~\ref{eq:intervention}, so $z'=\operatorname{Steer}(z,\alpha u)$ (Section~\ref{sec:activationsteering}). Control is enabled for $\alpha>0$ and disabled for $\alpha=0$, and applies to all valid prefill and decode tokens in target and protection requests.

\paragraph{Compute on- and off-state supervision.}
Let $D_A$, $D_B$, and $D_G$ denote the datasets of target examples, protection examples, and generic replay text, respectively. With control enabled, target supervision rewards the intended behavior on $D_A$, while Kullback--Leibler divergence (KL) from the teacher encourages preservation on $D_B$ under the same intervention. With control disabled, teacher KL encourages fidelity on $D_A$, $D_B$, and $D_G$. These losses are evaluated on model outputs.

Target supervision specifies the behavior that the intervention should strengthen. Let $i$ index examples in $D_A$, and let $\ell_A(M,\alpha v;i)$ denote a differentiable target loss for example $i$ under control $\alpha v$, with lower values indicating better target control. Jointly minimizing this loss with respect to the replacement parameters and auxiliary vector encourages the intervened model to exhibit behavior $A$. Appendix~\ref{sec:supervision} specifies the loss used for the Corrigibility task in our experiments.

For fidelity, let $D$ denote a dataset and $x$ one of its examples. Let $\mathcal T(x)$ be the set of evaluated token positions, $t$ a position in that set, and $c_t(x)$ the fixed context preceding that token: the prompt and teacher-generated prefix for behavioral requests, or the preceding source text for replay. Write $p_{M_0}(\cdot\mid c_t(x))$ for the unsteered teacher's next-token distribution and $p_{M,\alpha v}(\cdot\mid c_t(x))$ for the corresponding distribution of the evaluated model under control. Both distributions range over the full vocabulary. We denote empirical averaging over examples by $\mathbb E_{x\sim D}$ and the number of evaluated positions by $|\mathcal T(x)|$. The fidelity loss $\mathcal K_D$ is their average forward KL:
\begin{equation}
\mathcal K_D(M,\alpha v)=\mathbb E_{x\sim D}
\left[\frac{1}{|\mathcal T(x)|}\sum_{t\in\mathcal T(x)}
\KL\!\left(p_{M_0}(\cdot\mid c_t(x))\,\|\,p_{M,\alpha v}(\cdot\mid c_t(x))\right)\right].
\label{eq:kl}
\end{equation}
Averaging first within a request and then across requests prevents longer responses from receiving proportionally more weight. Figure~\ref{fig:workflow} abbreviates the replacement's output distribution as $p_{M_\theta}$, with the control state implicit. Teacher responses on $A$ and $B$ contain at most 128 tokens, include an end-of-sequence (EOS) token only when actually generated, and receive no artificial EOS after truncation. Replay uses next-token positions in the source text.

Let $\mathcal L_A^{\mathrm{on}}$ and $\mathcal L_B^{\mathrm{on}}$ denote target and protection losses under enabled control, and $\mathcal L_{\mathrm{off}}$ fidelity loss without control. With $\mathbb E_{i\sim D_A}$ denoting the mean over target items, the joint objective is $\mathcal L(\theta,u)$:
\begin{align}
\mathcal L(\theta,u)&=\mathcal L_A^{\mathrm{on}}+\mathcal L_B^{\mathrm{on}}+\mathcal L_{\mathrm{off}},\label{eq:loss}\\
\mathcal L_A^{\mathrm{on}}&=\mathbb E_{i\sim D_A}[\ell_A(M_\theta,\alpha u;i)],
\qquad \mathcal L_B^{\mathrm{on}}=\mathcal K_{D_B}(M_\theta,\alpha u),\nonumber\\
\mathcal L_{\mathrm{off}}&=\frac{1}{3}\sum_{D\in\{D_A,D_B,D_G\}}\mathcal K_D(M_\theta,0).\nonumber
\end{align}
The target term preserves controllability, the protection term penalizes changes to the original model's behavior on $B$ under steering, and the off term preserves unsteered outputs. Protection supervision shapes the replacement before controller fitting. All three loss coefficients are one, and the off distributions are weighted equally.

\paragraph{Jointly optimize the replacement and auxiliary vector.}
We update the encoder, decoder, their biases, and $u$ using Equation~\ref{eq:loss}, with gradients propagating through the frozen downstream computation. The activation settings remain fixed: the thresholds for JumpReLU and the number of retained features for TopK. After training, we measure $L_0$ to track latent activation.

\subsection{Using steering methods on the replacement}
\label{sec:deployment}
Following Figure~\ref{fig:workflow}(B), we freeze the replacement, fit a direction, and apply steering.

\paragraph{Freeze the replacement and set aside the auxiliary vector.}
After training, we freeze $\theta$ and set aside $u$. The resulting $M_\theta$ supplies latent representations and frozen computation for the downstream method, whose final steering vector is fitted separately from $u$.

\paragraph{Fit the selected steering method in the latent space.}
For method $j$ and its fitting data $\mathcal D_j$, we apply the rule defined in Equation~\ref{eq:steeringmethod} to the replacement model:
\begin{equation}
v_j=\operatorname{Fit}_j(M_\theta,\mathcal D_j),
\qquad v_{j,\bar I}=0.
\label{eq:fitreplacement}
\end{equation}
Here the chosen representation site is the transcoder's latent space, with the fixed intervention support $I$ from stage (A). Methods based on representation statistics compute them on $M_\theta$; optimization-based methods evaluate their objectives through its frozen computation. The selected method retains its fitting rule and objective, with $\theta$ fixed. For optimization-based fitting, gradients reach $v_j$ through the frozen decoder and downstream computation required by the objective. Section~\ref{sec:experiments} and Appendix~\ref{sec:implementation} give the evaluated methods and fitting configurations.

\paragraph{Set the strength and apply steering.}
Let $b$ be the nominal perturbation norm used for dense baseline directions, $W_{\mathrm{dec}}[I]$ the decoder rows corresponding to the support $I$, and $\|\cdot\|_2$ the Euclidean norm. We normalize the latent direction so that $\|v_{j,I}W_{\mathrm{dec}}[I]\|_2=b$, matching that norm in the decoded output space. The scalar $\alpha$ sets the intervention strength; our experiments fix it at two for all steered conditions (Section~\ref{sec:setup}). For the JumpReLU case, clipping in Equation~\ref{eq:intervention} makes the realized perturbation input-dependent. For encoded latents $z$, we apply
\begin{equation}
z'=\operatorname{Steer}(z,\alpha v_j),
\qquad y'=z'W_{\mathrm{dec}}+b_{\mathrm{dec}}.
\label{eq:applyreplacement}
\end{equation}
Here $\operatorname{Steer}$ uses the same activation-specific rule from Equation~\ref{eq:intervention} as in stage (A). The decoded $y'$ replaces the MLP output without a teacher reconstruction error. Control acts on all valid prefill and decode tokens in target and protection requests. Setting $\alpha=0$ gives the unsteered $M_\theta$-off model. Dense baselines add directions at the corresponding block output in $M_0$.

\section{Empirical Evaluation}
\label{sec:experiments}

Our experiments test whether \method\ reduces safety side effects while preserving target control, and whether protection learned from one safety dataset generalizes to unseen datasets. We compare three steering methods on the original and replacement versions of three language models, emphasizing methods that have no protection objective of their own. Section~\ref{sec:setup} describes the tasks, datasets, and models, and Section~\ref{sec:metrics} defines the evaluation metrics. Section~\ref{sec:saladresults} examines the SALAD-Bench development split, which shares its dataset source with protection training, and Section~\ref{sec:oodresults} evaluates transfer to three safety datasets excluded from training and development.

\subsection{Experimental setup}
\label{sec:setup}

\paragraph{Tasks and datasets.}
We evaluate target control through Corrigibility preferences: willingness to accept correction or modification. Each item contains a prompt and two answer options expressing the target and contrasting preferences. Target training examples come from the CAA/OPIUM task assets \citep{rimsky2024caa,aravindan2026opium}; evaluation uses 201 development and 716 test items from Anthropic's model-written evaluations \citep{perez2022modelwritten}.

To measure safety side effects, we generate free-form responses to harmful requests and assess whether they satisfy the harmful intent. SALAD-Bench \citep{li2024saladbench} supplies all protection training data and a separate development split of 256 requests in 239 intent clusters. The unseen test set pools 240 HarmBench \citep{mazeika2024harmbench}, 520 AdvBench \citep{zou2023universal}, and 313 StrongREJECT \citep{souly2024strongreject} requests, totaling 1,073 requests in 481 intent clusters. These three datasets are used exclusively for testing in all models. We use ``in-distribution'' for the SALAD development evaluation and ``out-of-distribution'' for the pooled unseen safety datasets; each is paired with the corresponding Corrigibility split. Alpaca \citep{taori2023alpaca} provides general training text.

\paragraph{Models and evaluation conditions.}
We evaluate Gemma-3-4B-it and Gemma-3-12B-it \citep{gemmateam2025gemma3}, and Llama-3.1-8B-Instruct \citep{grattafiori2024llama3}. For each base model, we compare CAA \citep{rimsky2024caa}, target-only SSV, and OPIUM \citep{aravindan2026opium} on the original model $M_0$ and its STR replacement $M_\theta$, together with their unsteered (off) baselines. We fix $\alpha=2$ across all steered conditions to compare side-effect reduction under a common nominal intervention budget within each base model. This multiplier has also been used for Llama-3.1-8B-Instruct in prior work on steering externalities \citep{aravindan2026opium}. Appendix~\ref{sec:implementation} provides training and implementation details.

\subsection{Evaluation metrics}
\label{sec:metrics}

\paragraph{Target utility ($U$).}
\label{sec:targetutility}
Target utility measures the probability assigned to the target-aligned answer, normalized over the two options. Let $i$ index evaluation items and $o\in\{1,2\}$ index the original and swapped option orders. Let $x_i^{(o)}$ be the prompt under order $o$, and $y_{i,o}^{+}$ and $y_{i,o}^{-}$ the target-aligned and contrasting answers. Write $P_{M,\alpha v}(y\mid x)$ for the probability that model $M$, under control $\alpha v$, assigns to answer $y$ given prompt $x$. We average over both orders to reduce position sensitivity:
\begin{equation}
U_i(M,\alpha v)=\frac{1}{2}\sum_{o=1}^{2}
\frac{P_{M,\alpha v}(y_{i,o}^{+}\mid x_i^{(o)})}
{P_{M,\alpha v}(y_{i,o}^{+}\mid x_i^{(o)})+
 P_{M,\alpha v}(y_{i,o}^{-}\mid x_i^{(o)})}.
\label{eq:utility}
\end{equation}
The dataset score $U$ averages $U_i$ over items; higher values indicate stronger target preference.

\paragraph{Attack success rate (ASR).}
Safety degradation $E_B$ is measured by harmful-request ASR, using a fixed Qwen3-8B judge \citep{yang2025qwen3} and scoring rubric. For each request, we first average attack-success judgments over the evaluated runs, then average requests within each intent cluster, and finally average the cluster scores with equal weight. The pooled test ASR therefore weights intent clusters equally across the three datasets. Lower ASR indicates better preservation of safety.

\paragraph{Changes relative to the original model ($\Delta U$, $\Delta\ASR$).}
For both original-model steering and STR, target gains and safety side effects use the same original unsteered reference, $M_0$-off:
\begin{equation}
\Delta U(M,v)=U(M,\alpha v)-U(M_0,0),
\label{eq:targetgain}
\end{equation}
\begin{equation}
\Delta E_B(M,v)=E_B(M,\alpha v)-E_B(M_0,0).
\label{eq:externality}
\end{equation}
Here $M\in\{M_0,M_\theta\}$ and $\Delta E_B=\Delta\ASR$. Positive $\Delta U$ indicates a target gain, while positive $\Delta\ASR$ indicates a safety side effect. Scores and additive differences use \% notation. We report 95\% confidence intervals (CIs) to quantify uncertainty in paired STR-minus-original differences.

\subsection{In-distribution results on SALAD-Bench}
\label{sec:saladresults}
\begin{table}[t]
\centering
\caption{In-distribution evaluation on the SALAD-Bench development split.}
\label{tab:salad}
\begin{tabular}{@{}lrrrrrr@{}}
\toprule
& \multicolumn{2}{c}{Gemma-3-4B} & \multicolumn{2}{c}{Llama-3.1-8B} & \multicolumn{2}{c}{Gemma-3-12B} \\
\cmidrule(lr){2-3}\cmidrule(lr){4-5}\cmidrule(l){6-7}
Condition & $U\uparrow$ & ASR$\downarrow$ & $U\uparrow$ & ASR$\downarrow$ & $U\uparrow$ & ASR$\downarrow$ \\
\midrule
$M_0$-off & 78.96 & 4.60 & 79.60 & 2.44 & 77.98 & 3.07 \\
$M_\theta$-off & 80.02 & 4.30 & 82.55 & 2.74 & 81.44 & 2.98 \\
\midrule
$M_0$-CAA & 81.46 & 32.22 & 86.06 & 6.97 & 84.13 & 26.71 \\
$M_\theta$-CAA & 81.09 & 4.95 & 86.10 & 2.37 & 87.49 & 4.07 \\
\midrule
$M_0$-SSV & 80.69 & 36.12 & 77.39 & 4.97 & 84.74 & 29.29 \\
$M_\theta$-SSV & 81.17 & 5.72 & 88.87 & 2.93 & 87.51 & 3.63 \\
\midrule
$M_0$-OPIUM & 81.95 & 29.96 & 83.50 & 2.58 & 84.13 & 22.78 \\
$M_\theta$-OPIUM & 80.97 & 5.23 & 83.64 & 2.49 & 87.47 & 3.42 \\
\bottomrule
\end{tabular}
\par\smallskip
\begin{minipage}{\linewidth}
\normalsize
All steered conditions use $\alpha=2$. Target utility $U$ is evaluated separately on 201 Corrigibility development items; ASR uses 256 SALAD requests in 239 intent clusters. All values are percentages; higher $U$ and lower ASR are better. Off denotes no activation steering. $M_0$ is the original model and $M_\theta$ the STR replacement. Each original/STR pair uses the same steering method.
\end{minipage}
\end{table}

Table~\ref{tab:salad} reports held-out development results for SALAD safety and Corrigibility utility.

STR lowers ASR for CAA and target-only SSV across all three models. For SSV, ASR falls from 36.12\% to 5.72\% on Gemma-3-4B and from 29.29\% to 3.63\% on Gemma-3-12B, with higher target utility in both cases. On Llama-3.1-8B, STR raises SSV utility from 77.39\% to 88.87\%, reversing the decline below the original unsteered baseline, while also reducing ASR. CAA retains nearly unchanged utility on Gemma-3-4B and Llama and improves it on Gemma-3-12B, with STR ASRs close to the original unsteered levels. Every STR controller gains utility over $M_0$-off, while CAA/SSV keep ASR much closer to this reference than original-model steering.

OPIUM also shows substantial ASR reductions on both Gemma models, with a modest utility decrease on Gemma-3-4B and an increase on Gemma-3-12B. On Llama, both metrics remain close to original-model OPIUM. Overall, replacement training benefits target-focused steering without changing its fitting objective; we next test transfer beyond SALAD.

\subsection{Out-of-distribution results on unseen safety datasets}
\label{sec:oodresults}
\begin{table}[t]
\centering
\caption{Out-of-distribution evaluation.}
\label{tab:ood}
\begin{tabular}{@{}lrrrrrr@{}}
\toprule
& \multicolumn{2}{c}{Gemma-3-4B} & \multicolumn{2}{c}{Llama-3.1-8B} & \multicolumn{2}{c}{Gemma-3-12B} \\
\cmidrule(lr){2-3}\cmidrule(lr){4-5}\cmidrule(l){6-7}
Condition & $U\uparrow$ & ASR$\downarrow$ & $U\uparrow$ & ASR$\downarrow$ & $U\uparrow$ & ASR$\downarrow$ \\
\midrule
$M_0$-off & 77.11 & 11.44 & 76.71 & 17.16 & 73.15 & 10.92 \\
$M_\theta$-off & 78.50 & 10.85 & 79.25 & 16.29 & 78.19 & 11.09 \\
\midrule
$M_0$-CAA & 80.31 & 45.85 & 83.15 & 25.82 & 81.46 & 32.77 \\
$M_\theta$-CAA & 79.53 & 14.49 & 82.23 & 14.41 & 84.13 & 12.29 \\
\midrule
$M_0$-SSV & 79.74 & 42.46 & 74.10 & 20.37 & 82.02 & 34.97 \\
$M_\theta$-SSV & 79.57 & 14.42 & 84.73 & 15.36 & 84.13 & 12.91 \\
\midrule
$M_0$-OPIUM & 80.66 & 38.85 & 80.75 & 17.41 & 81.35 & 26.33 \\
$M_\theta$-OPIUM & 79.35 & 13.83 & 80.18 & 16.54 & 84.09 & 12.36 \\
\bottomrule
\end{tabular}
\par\smallskip
\begin{minipage}{\linewidth}
\normalsize
All steered conditions use $\alpha=2$. Target utility $U$ uses 716 Corrigibility test items. ASR uses the pooled 1,073 requests from HarmBench, AdvBench, and StrongREJECT, with ASR weighting the 481 intent clusters equally. All values are percentages; higher $U$ and lower ASR are better. Off denotes no activation steering. $M_0$ is the original model and $M_\theta$ the STR replacement. Each original/STR pair uses the same steering method.
\end{minipage}
\end{table}

\begin{table}[t]
\centering
\caption{Paired test differences between STR and original-model steering with the same method.}
\label{tab:paired}
\begin{tabular}{@{}llrr@{}}
\toprule
Model & Method & $\Delta U_{\mathrm{STR}}-\Delta U_{\mathrm{original}}$ & $\Delta\ASR_{\mathrm{STR}}-\Delta\ASR_{\mathrm{original}}$ \\
\midrule
Gemma-3-4B & CAA & $-0.77\;[-2.14, +0.61]$ & $-31.36\;[-35.81, -26.91]$ \\
 & SSV & $-0.16\;[-1.50, +1.17]$ & $-28.04\;[-31.93, -24.21]$ \\
 & OPIUM & $-1.32\;[-2.69, +0.11]$ & $-25.02\;[-28.71, -21.45]$ \\
\midrule
Llama-3.1-8B & CAA & $-0.91\;[-1.52, -0.30]$ & $-11.41\;[-14.30, -8.67]$ \\
 & SSV & $+10.64\;[+9.15, +12.13]$ & $-5.00\;[-7.20, -2.91]$ \\
 & OPIUM & $-0.57\;[-1.07, -0.05]$ & $-0.86\;[-2.79, +1.02]$ \\
\midrule
Gemma-3-12B & CAA & $+2.67\;[+1.46, +3.93]$ & $-20.48\;[-24.16, -16.88]$ \\
 & SSV & $+2.12\;[+0.93, +3.36]$ & $-22.06\;[-25.49, -18.79]$ \\
 & OPIUM & $+2.74\;[+1.54, +3.99]$ & $-13.98\;[-16.96, -11.05]$ \\
\bottomrule
\end{tabular}
\par\smallskip
\begin{minipage}{\linewidth}
\normalsize
Subscripts STR and original denote steering on $M_\theta$ and $M_0$ at $\alpha=2$. The common $M_0$-off reference in each $\Delta$ cancels in these paired differences. Brackets give 95\% confidence intervals; all values use \%. Positive utility and negative ASR differences favor STR. Unrounded estimates can differ from subtraction of displayed means.
\end{minipage}
\end{table}

Table~\ref{tab:ood} reports pooled test results on HarmBench, AdvBench, and StrongREJECT, all excluded from protection training and development, together with held-out Corrigibility utility. Table~\ref{tab:paired} gives paired comparisons; Appendix~\ref{sec:datasetresults} gives request-averaged ASR by benchmark and utility by source.

SSV shows the most consistent transfer. On Gemma-3-4B, STR lowers ASR from 42.46\% to 14.42\%, while the observed mean utilities are close (79.74\% versus 79.57\%). On Gemma-3-12B, ASR falls from 34.97\% to 12.91\% and utility improves. On Llama-3.1-8B, utility rises from 74.10\% to 84.73\% while ASR decreases from 20.37\% to 15.36\%. The paired intervals support ASR reductions in all three models and utility gains in the latter two. Against the common original unsteered reference, STR--SSV retains positive target gains in every model, with smaller ASR increases on Gemma and ASR below baseline on Llama.

CAA also reduces ASR across all three models, with every paired ASR interval below zero. Utility improves on Gemma-3-12B but decreases modestly on the other two models; the decrease on Llama is supported by its paired CI. For OPIUM, STR substantially reduces ASR on both Gemma models, improving utility on Gemma-3-12B while its small decrease on Gemma-3-4B has a CI spanning zero. On Llama, the estimated ASR reduction is small and its paired CI includes zero, alongside a small utility decrease.

These results extend the protection observed on SALAD to unseen safety datasets, particularly for CAA and SSV, whose fitting objectives contain no protection term. All STR controllers retain positive target gains relative to the original unsteered models (Table~\ref{tab:ood}), although the utility trade-off against original-model steering varies by method and model.

\FloatBarrier
\section{Related Work}
\label{sec:related}

\method\ aims to preserve non-target behavior while enabling effective activation steering. This goal connects our work to three research directions: understanding steering side effects and behavioral specificity, preserving safety during steering, and maintaining generation quality through selective interventions. We relate each direction to supervised replacement training.

\paragraph{Steering side effects and specificity.}
Activation steering enables targeted behavioral changes through interventions on internal representations, as demonstrated by ActAdd and CAA \citep{turner2023activation,rimsky2024caa}. Subsequent work examines effects beyond the steering objective. \citet{xiong2026externalities} show that steering with benign compliance or formatting data can increase jailbreak vulnerability. \citet{korznikov2025rogue} further find that steering benign SAE features can weaken safety safeguards. \citet{goyal2026specificity} distinguish general, control, and robustness specificity: in their experiments, steering can preserve general abilities and standard refusal behavior while weakening robustness to adversarial prompts. Across a broader taxonomy, \citet{ong2026forecasting} characterize and forecast cross-behavior effects spanning 67 behaviors and three models. These findings establish why target efficacy and non-target preservation must be evaluated separately.

\paragraph{Preserving safety during behavioral steering.}
OPIUM \citep{aravindan2026opium} directly addresses the same goal as \method: retaining a desired intervention while suppressing harmful side effects. It optimizes a steering vector by matching downstream representations to a target-steered reference on utility prompts and a safer reference on protection prompts. Its evaluations cover safety degradation from utility steering and over-refusal from refusal steering. \method\ places target and protection supervision in the training of a transcoder that replaces an MLP computation. Steering methods then fit directions on this frozen replacement.

\paragraph{Preserving generation quality through selective steering.}
Other approaches reduce unintended changes by making steering vectors more selective. SAE-Targeted Steering (SAE-TS) \citep{chalnev2024saets} learns a linear model of how steering vectors affect SAE features and uses it to construct vectors designed to increase a chosen feature with fewer unintended changes. SAE-SSV \citep{he2025saessv} selects a task-relevant SAE subspace and learns a vector using target, language-modeling, and sparsity losses to balance control and generation quality. These methods improve the intervention within an existing representation space. \method\ learns the replacement mapping itself, using supervision on specified non-target behavior to reduce its degradation under steering.

\section{Conclusion}
\label{sec:conclusion}

We proposed \method\ to reduce the impact of model steering on non-target behavior. Supervised transcoder training learns a replacement for the MLP computation at the steering layer; selected steering methods then fit directions on this frozen replacement while retaining their original objectives. Across three steering methods and three language models, the experiments show that a replacement trained with SALAD protection supervision can reduce safety side effects on unseen benchmark sources while retaining effective target control. Target-only SSV shows the most consistent improvement, maintaining similar observed utility on Gemma-3-4B and improving both utility and safety on Llama-3.1-8B and Gemma-3-12B. These findings support learning preservation in the computation receiving the intervention, benefiting steering methods without protection objectives.

\clearpage
\subsection*{AI use statement}
AI assistants were used solely to polish the language of the manuscript and were not involved in method design or implementation, or in designing or conducting experiments.

\subsection*{Reproducibility statement}
Section~\ref{sec:method} specifies the replacement training objective and steering procedure, Section~\ref{sec:experiments} describes the evaluation setup and metrics, and Appendix~\ref{sec:implementation} records model integration, controller fitting, and inference configurations. We will provide anonymized training and evaluation code, together with experiment configurations, as supplementary material to support reproduction of the reported results.

\subsection*{Ethics statement}
This work aims to reduce safety degradation caused by model steering. The evaluation uses existing harmful-request benchmarks to measure whether models comply with harmful instructions. Such requests and generated responses may contain harmful content, and steering techniques can themselves be misused to weaken safeguards. The reported reductions in attack success rate characterize the evaluated settings and do not establish safety for unrestricted deployment. Any reuse or redistribution of the models, datasets, and code should follow their applicable licenses and access conditions, with care to avoid disseminating actionable harmful outputs.

\clearpage
\bibliography{references/references}

\clearpage
\appendix
\section{Implementation Details}
\label{sec:implementation}

\subsection{Model integration and supervision}
\label{sec:supervision}
Each replacement substitutes one MLP mapping and preserves the surrounding residual connection. The Gemma-3-4B and Gemma-3-12B replacements use 65,536 JumpReLU features at layers 17 and 24, respectively. The Gemma mapping connects the pre-feedforward normalization output to the post-feedforward normalization output, so the latter normalization is not applied a second time. Llama-3.1-8B uses a 131,072-feature TopK transcoder at layer 15 with $k=128$. Signed latent intervention follows encoding without a second TopK operation. Layers are zero-indexed.

The intervention support contains $K=256$ coordinates, distinct from the TopK activation count $k$, selected before supervised training using standardized positive--negative contrasts in the transcoder's latent representations of the target training data. The support remains fixed throughout replacement training and subsequent controller fitting. We jointly optimize the encoder, decoder, biases, and auxiliary vector while freezing the surrounding model and activation settings. The three terms in Equation~\ref{eq:loss} have equal coefficients. Generic replay uses Alpaca \citep{taori2023alpaca}; target supervision uses Corrigibility task assets \citep{rimsky2024caa,aravindan2026opium}. SALAD-Bench \citep{li2024saladbench} supplies all safety protection training data for each model.

For Corrigibility training examples, the target loss is $\ell_A(M,\alpha v;i)=-\log U_i(M,\alpha v)$, with $U_i$ defined in Equation~\ref{eq:utility}. The logarithm is taken after averaging the normalized answer probability over the two option orders. Replacement training uses $D_A$, $M=M_\theta$, and $v=u$.

\subsection{Fitting and applying the controllers}
CAA \citep{rimsky2024caa} extracts a positive--negative contrast using the last content token of the target responses. SSV starts from the CAA direction and optimizes only the target loss $-\log U_i$. OPIUM \citep{aravindan2026opium} also starts from CAA, but optimizes both target and protection terms. Original-model controllers are fitted on $M_0$; replacement controllers use the frozen replacement's latent space. The training auxiliary vector is not deployed as a controller.

For OPIUM, let $H_{M,v}$ be the dense hidden state at the matching layer's last valid prompt token for model $M\in\{M_0,M_\theta\}$ under candidate direction $v$ at unit strength. The matching layer is 21 for Gemma-3-4B, 19 for Llama-3.1-8B, and 28 for Gemma-3-12B. Let $H_{M,\mathrm{CAA}}$ and $H_{M,0}$ denote cached, detached reference states under that model's CAA direction and without steering, respectively. The dependence on the input prompt is implicit. With $\operatorname{MSE}$ denoting mean squared error and $\mathbb E_A$ and $\mathbb E_B$ denoting empirical averages over the target and protection training data, the objective is
\begin{equation}
\mathcal L_{\mathrm{OPIUM}}=
\mathbb E_A\operatorname{MSE}(H_{M,v},H_{M,\mathrm{CAA}})
+15\mathbb E_B\operatorname{MSE}(H_{M,v},H_{M,0}).
\label{eq:opium}
\end{equation}
These references differ from the original teacher used to train the replacement. On $M_\theta$, only the latent vector is optimized, through the frozen decoder and downstream computation.

We use seeds 42, 43, and 44 for replacement training and controller fitting; original-model CAA is deterministic.

\subsection{Inference configuration}
Fitted directions use the normalization in Section~\ref{sec:deployment} and are scaled by $\alpha=2$. For latent directions, normalization sets the nominal perturbation norm in the decoded output space; JumpReLU clipping can make the realized perturbation input-dependent. Control applies to all valid prefill and decode tokens, using the same rule on target and safety requests without routing by task identity or harmfulness. The off conditions retain their respective model computations but disable activation steering. Safety responses are generated with a limit of 2,048 new tokens.

\clearpage
\section{Test Results by Dataset and Source}
\label{sec:datasetresults}

We break down the test results of Section~\ref{sec:oodresults} across individual safety benchmarks and target-item sources. These results use the same completed evaluations at $\alpha=2$ and report point estimates.

\subsection{Safety results by benchmark}
\label{sec:benchmarkasr}

Table~\ref{tab:benchmarkasr} reports ASR separately for HarmBench \citep{mazeika2024harmbench}, AdvBench \citep{zou2023universal}, and StrongREJECT \citep{souly2024strongreject}. Each score averages requests equally within the dataset, then averages across evaluated runs. The pooled ASR in Table~\ref{tab:ood} weights intent clusters equally and cannot be recovered by averaging benchmark scores, either equally or by sample count.

\begin{table}[htbp]
\centering
\caption{Request-averaged attack success rates on unseen safety benchmarks.}
\label{tab:benchmarkasr}
\begin{tabular}{@{}lrrr@{}}
\toprule
Condition & \shortstack{HarmBench\\240 requests} & \shortstack{AdvBench\\520 requests} & \shortstack{StrongREJECT\\313 requests} \\
\midrule
\multicolumn{4}{l}{\textbf{Gemma-3-4B-it}} \\
$M_0$-off & 25.42 & 0.96 & 1.28 \\
$M_\theta$-off & 22.64 & 1.54 & 2.34 \\
$M_0$-CAA & 57.08 & 35.58 & 37.06 \\
$M_\theta$-CAA & 29.58 & 1.79 & 2.98 \\
$M_0$-SSV & 53.89 & 33.27 & 34.08 \\
$M_\theta$-SSV & 29.72 & 1.99 & 2.34 \\
$M_0$-OPIUM & 50.83 & 22.31 & 29.39 \\
$M_\theta$-OPIUM & 28.75 & 1.79 & 2.45 \\
\midrule
\multicolumn{4}{l}{\textbf{Llama-3.1-8B-Instruct}} \\
$M_0$-off & 28.33 & 5.77 & 3.51 \\
$M_\theta$-off & 27.78 & 6.03 & 3.51 \\
$M_0$-CAA & 40.00 & 13.65 & 7.67 \\
$M_\theta$-CAA & 25.42 & 4.94 & 3.30 \\
$M_0$-SSV & 32.22 & 8.33 & 5.43 \\
$M_\theta$-SSV & 26.94 & 4.81 & 3.73 \\
$M_0$-OPIUM & 29.17 & 6.09 & 3.41 \\
$M_\theta$-OPIUM & 28.06 & 6.99 & 3.30 \\
\midrule
\multicolumn{4}{l}{\textbf{Gemma-3-12B-it}} \\
$M_0$-off & 25.00 & 1.73 & 0.32 \\
$M_\theta$-off & 26.53 & 1.09 & 0.11 \\
$M_0$-CAA & 47.92 & 10.77 & 19.81 \\
$M_\theta$-CAA & 27.22 & 1.22 & 0.53 \\
$M_0$-SSV & 49.86 & 12.56 & 21.73 \\
$M_\theta$-SSV & 27.92 & 1.54 & 0.64 \\
$M_0$-OPIUM & 41.53 & 8.08 & 12.35 \\
$M_\theta$-OPIUM & 27.36 & 1.15 & 0.43 \\
\bottomrule
\end{tabular}
\par\smallskip
\begin{minipage}{\linewidth}
\normalsize
ASR is reported in \%; lower is better. All steered conditions use $\alpha=2$. $M_0$ denotes the original model, $M_\theta$ the STR replacement, and off disables activation steering. Each original/STR pair uses the same steering method. Appendix~\ref{sec:benchmarkasr} details the aggregation of these benchmark point estimates.
\end{minipage}
\end{table}

CAA and SSV have lower ASR on the STR replacement for every benchmark and model. HarmBench retains the highest absolute ASR among the three benchmarks in all STR conditions. OPIUM also lowers ASR on every benchmark for both Gemma models. On Llama, its HarmBench and StrongREJECT scores decrease slightly, while AdvBench ASR rises from 6.09\% to 6.99\%. Target-focused steering benefits consistently, with gains varying across methods and datasets.

\clearpage
\subsection{Target utility by source}
\label{sec:sourceutility}

Table~\ref{tab:sourceutility} partitions the 716 Corrigibility test items \citep{perez2022modelwritten} into four source subsets: less-HHH, neutral-HHH, more-HHH, and no-goal-change. Neutral-HHH corresponds to the \texttt{anthropic\_original\_lm} source, and no-goal-change to \texttt{persona\_no\_goal\_change}. The no-goal-change estimate uses only 24 items. Each entry is the mean target-option utility $U$ defined in Equation~\ref{eq:utility}; these items are separate from the harmful-request benchmarks. Weighting the four source scores by their item counts recovers the overall test utility in Table~\ref{tab:ood}, up to rounding.

\begin{table}[htbp]
\centering
\caption{Target utility by Corrigibility test-item source.}
\label{tab:sourceutility}
\begin{tabular}{@{}lrrrr@{}}
\toprule
Condition & \shortstack{less-HHH\\252 items} & \shortstack{neutral-HHH\\138 items} & \shortstack{more-HHH\\302 items} & \shortstack{no-goal-change\\24 items} \\
\midrule
\multicolumn{5}{l}{\textbf{Gemma-3-4B-it}} \\
$M_0$-off & 54.16 & 91.61 & 92.61 & 39.47 \\
$M_\theta$-off & 56.40 & 92.24 & 93.72 & 40.08 \\
$M_0$-CAA & 61.62 & 92.35 & 93.29 & 43.86 \\
$M_\theta$-CAA & 65.21 & 89.65 & 90.34 & 35.74 \\
$M_0$-SSV & 61.31 & 91.27 & 92.59 & 45.10 \\
$M_\theta$-SSV & 65.23 & 89.75 & 90.36 & 35.90 \\
$M_0$-OPIUM & 61.91 & 93.19 & 93.66 & 42.00 \\
$M_\theta$-OPIUM & 65.09 & 89.39 & 90.07 & 36.30 \\
\midrule
\multicolumn{5}{l}{\textbf{Llama-3.1-8B-Instruct}} \\
$M_0$-off & 53.94 & 90.66 & 92.04 & 42.60 \\
$M_\theta$-off & 58.78 & 92.15 & 92.79 & 49.57 \\
$M_0$-CAA & 64.61 & 95.47 & 96.50 & 38.90 \\
$M_\theta$-CAA & 63.52 & 94.54 & 94.85 & 49.25 \\
$M_0$-SSV & 66.34 & 82.18 & 78.97 & 47.82 \\
$M_\theta$-SSV & 70.65 & 94.61 & 94.62 & 51.37 \\
$M_0$-OPIUM & 59.40 & 94.52 & 95.69 & 37.88 \\
$M_\theta$-OPIUM & 59.94 & 93.09 & 93.71 & 48.21 \\
\midrule
\multicolumn{5}{l}{\textbf{Gemma-3-12B-it}} \\
$M_0$-off & 46.38 & 90.57 & 91.01 & 29.17 \\
$M_\theta$-off & 55.32 & 93.18 & 94.24 & 30.22 \\
$M_0$-CAA & 60.25 & 95.83 & 96.73 & 29.28 \\
$M_\theta$-CAA & 67.01 & 96.34 & 96.28 & 40.72 \\
$M_0$-SSV & 61.66 & 95.97 & 96.84 & 28.98 \\
$M_\theta$-SSV & 67.05 & 96.33 & 96.27 & 40.70 \\
$M_0$-OPIUM & 59.82 & 96.04 & 96.75 & 29.21 \\
$M_\theta$-OPIUM & 66.90 & 96.34 & 96.32 & 40.21 \\
\bottomrule
\end{tabular}
\par\smallskip
\begin{minipage}{\linewidth}
\normalsize
Target utility $U$ is reported in \%; higher is better. All steered conditions use $\alpha=2$. The four subsets total 716 items. $M_0$ and $M_\theta$ denote the original and STR models; off disables steering. Original/STR pairs use the same method. Entries are source-level point estimates.
\end{minipage}
\end{table}

On Gemma-3-4B, STR improves less-HHH utility for all three steering methods, with decreases on the other source subsets. On Gemma-3-12B, gains are concentrated in less-HHH and no-goal-change, while more-HHH utility decreases slightly. On Llama, STR--SSV improves utility on all four subsets; CAA and OPIUM have mixed source-level changes. These differences explain how the aggregate target score can combine improvements on some sources with declines on others.
\FloatBarrier

\end{document}